\documentclass{article}

\usepackage{microtype}
\usepackage{graphicx}
\usepackage{subcaption}
\usepackage{booktabs} 

\usepackage{hyperref}
\usepackage{xurl}

\usepackage[preprint]{icml2026}

\usepackage{amsmath}
\usepackage{amssymb}
\usepackage{mathtools}
\usepackage{amsthm}

\usepackage[capitalize,noabbrev]{cleveref}

\theoremstyle{plain}

\theoremstyle{definition}

\theoremstyle{remark}

\usepackage[textsize=tiny]{todonotes}

\icmltitlerunning{Dynamic sparsity in tree-structured feed-forward layers at scale}

\begin{document}

\twocolumn[
  \icmltitle{Dynamic sparsity in tree-structured feed-forward layers at scale}



  \icmlsetsymbol{equal}{*}

  \begin{icmlauthorlist}

    \icmlauthor{Reza Sedghi}{citec}
    \icmlauthor{Robin Schiewer}{citec}
    \icmlauthor{Anand Subramoney}{RUHL}
    \icmlauthor{David Kappel}{citec}
  \end{icmlauthorlist}

  \icmlaffiliation{citec}{CITEC, Bielefeld University, Germany }
  \icmlaffiliation{RUHL}{Department of Computer Science, Royal Holloway, University of London, UK }

  \icmlcorrespondingauthor{Reza Sedghi}{reza.sedghi@uni-bielefeld.de}
  \icmlcorrespondingauthor{Robin Schiewer}{robin.schiewer@uni-bielefeld.de}
  \icmlcorrespondingauthor{Anand Subramoney}{anand.subramoney@rhul.ac.uk}
  \icmlcorrespondingauthor{David Kappel}{david.kappel@uni-bielefeld.de}

  \icmlkeywords{Machine Learning, ICML}

  \vskip 0.3in
]



\printAffiliationsAndNotice{}  

\begin{abstract}
At typical context lengths, the feed-forward MLP block accounts for a large share of a transformer's compute budget, motivating sparse alternatives to dense MLP blocks.
We study sparse, tree-structured feed-forward layers as drop-in replacements for MLP blocks in deep transformer architectures, enabling conditional computation via hard hierarchical routing without a separate router network.
We demonstrate for the first time that this form of tree-structured conditional sparsity can be applied for  autoregressive language modeling and downstream question answering, including zero- and few-shot settings, and its scalability beyond 1B parameters.
Despite activating fewer than 5\% of the feed-forward block's units per token, our models match dense baselines under controlled training and fine-tuning protocols.
We further analyze training dynamics and identify an emergent auto-pruning effect: the interaction of hard routing with asymmetric nonlinearities progressively deactivates unused paths, yielding partial conversion of dynamic routing into static structural sparsity.
We show that simple architectural choices can modulate this behavior and recover balanced trees without auxiliary losses.
Overall, our work demonstrates that tree-structured feed-forward layers provide a scalable and controllable mechanism for sparsifying large transformer models.


\end{abstract}

\section{Introduction}

\begin{figure*}[ht!]
    \centering
    \begin{subfigure}{0.24\textwidth}
        \centering
        \includegraphics[width=\linewidth]{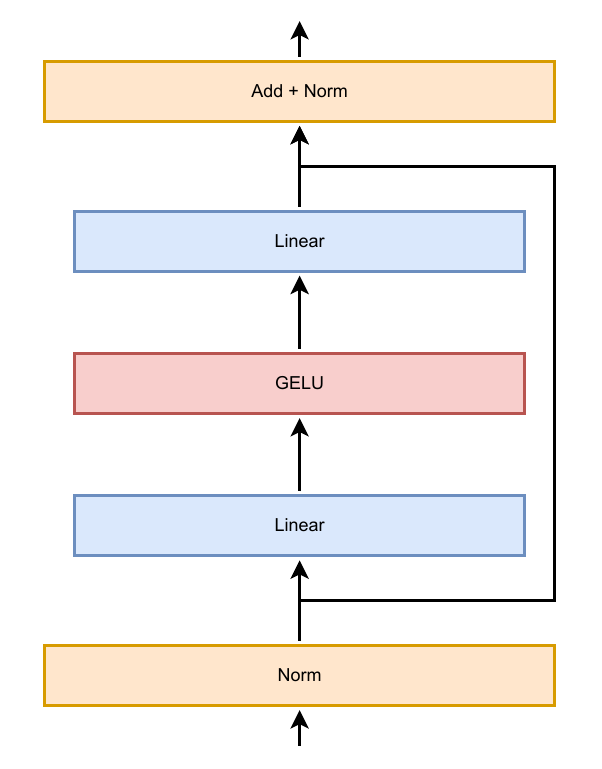}
        \caption{Standard MLP block}
    \end{subfigure}
    \hfill
    \begin{subfigure}{0.24\textwidth}
        \centering
        \includegraphics[width=\linewidth]{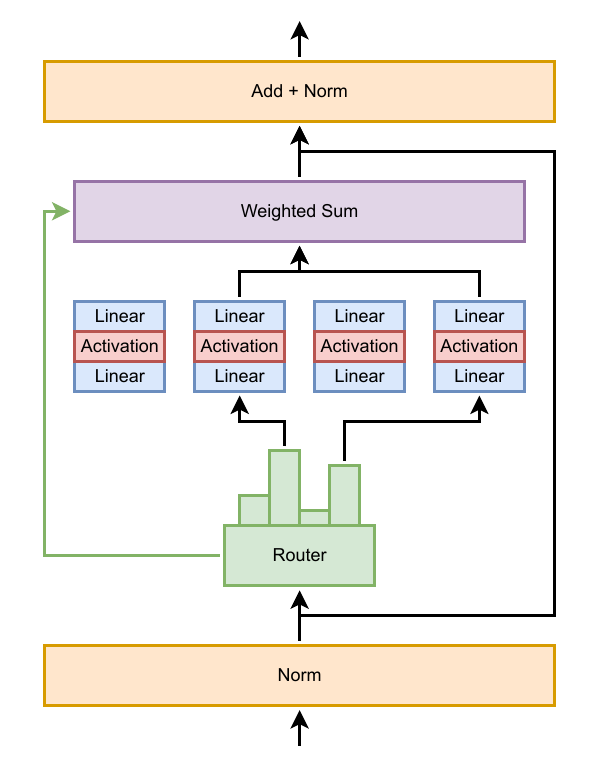}
        \caption{Mixture-of-Experts}
    \end{subfigure}
    \hfill
    \begin{subfigure}{0.465\textwidth}
        \centering
        \includegraphics[width=\linewidth]{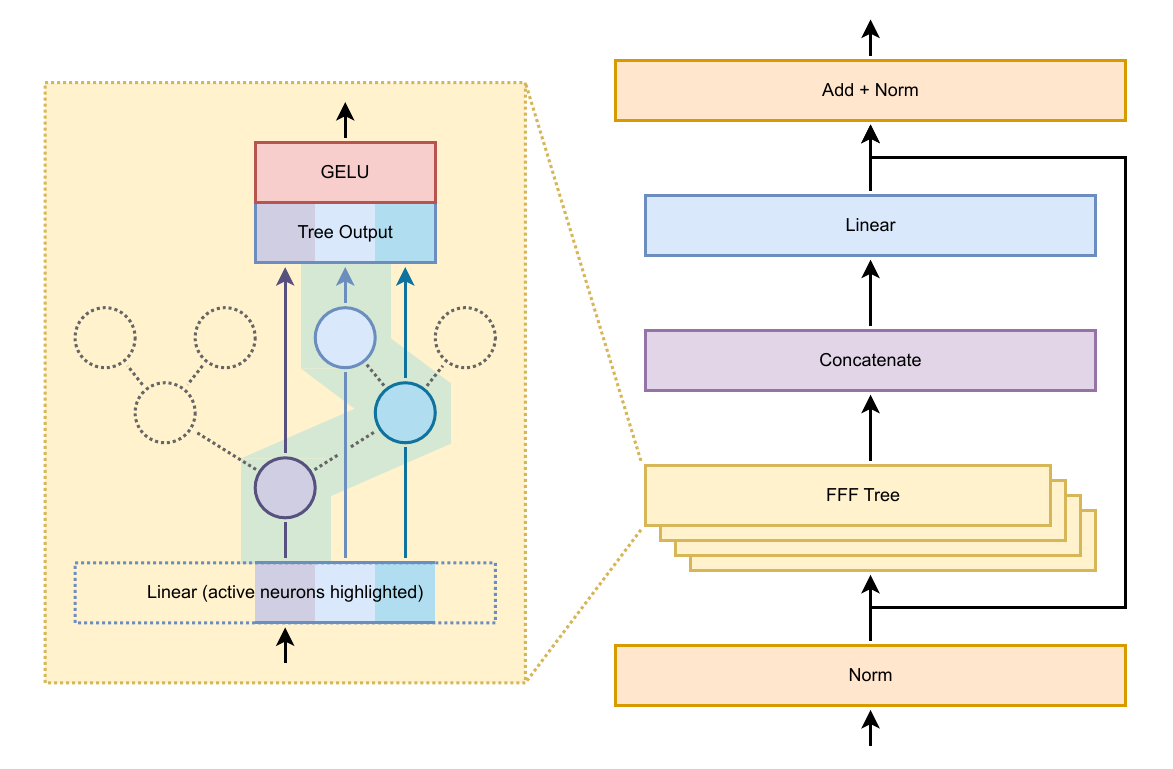}
        \caption{Fast FeedForward (FFF) layer}
    \end{subfigure}

    \caption{MLP layers and sparse replacements. \textbf{a)} The standard MLP feed-forward block. \textbf{b)} A top-k MoE block, featuring a router that recruits experts (linear-activation-linear sub-networks) on a per-token level. Here, $k = 2$ experts process each input token and their outputs are integrated via a weighted sum. \textbf{c)} An FFF block that contains $P=4$ parallel trees of depth $D=2$, shown in yellow. Tree output is integrated via a concatenation of the individual sparse tree outputs and a final linear projection. A single path through the tree (green) is activated through hard routing decisions to produce the layer output.}
    \label{fig:overview}
\end{figure*}

Feed-forward (FF) or MLP blocks account for a dominant fraction of parameters and computation in modern transformer architectures, particularly in large language models (LLMs). As models scale to billions of parameters, the cost of dense FF blocks has become a primary bottleneck, motivating extensive research into sparse alternatives. Among these, mixture-of-experts (MoE) models have emerged as a prominent approach to exploit dynamic sparsity by activating only a subset of parameters per token \cite{mu2025comprehensivesurveymixtureofexpertsalgorithms} (see Fig.~\ref{fig:overview}a,b). However, practical MoE systems typically rely on learned routing networks, auxiliary load-balancing losses, and specialized training and inference infrastructure \cite{lepikhin2020gshard, fedus2022switchtransformers, dai2022stablemoe, zhou2022mixtureofexpertswith, gale2023megablocks}.

Tree-structured conditional computation offers a different sparsification mechanism, that organizes computation hierarchically, such that each input activates a single root-to-leaf path in a binary tree. This induces \emph{predictable and structured sparsity}: routing decisions at each level deterministically select a subtree, allowing entire paths to be skipped without a separate router module or expert aggregation. This enables high levels of sparsity with minimal architectural overhead. Fast Feedforward (FFF) layers \cite{Belck2023FastFN, Belck2023ExponentiallyFL} instantiated this idea by replacing dense FF blocks by a sequence of hard binary routing decisions, activating only a small number of nodes while all other parameters remain inactive. As a result, FFF layers can achieve levels of dynamic sparsity exceeding $99\%$. However, apart from encoder-style BERT models for classification with ca. 300M parameters \cite{Belck2023ExponentiallyFL}, the capabilities and limitations of tree-structured conditional computation remains largely unexplored compared to widely used MoEs.




In this work, we address these open questions through a systematic empirical and analytical study of FFF layers in GPT-style transformers. We evaluate models trained from scratch and via fine-tuning across a wide range of scales, including models with over a billion parameters, and assess performance on language modeling as well as zero- and few-shot downstream benchmarks. Across these settings, we show that FFF-based models closely match dense feed-forward baselines while operating in highly sparse computational regimes.
Beyond scalability and performance, we discover a new 
emergent \emph{auto-pruning} effect, in which the interaction between hard routing decisions and asymmetric nonlinearities progressively deactivates entire tree paths during training. This process partially converts dynamic, input-dependent routing into stable structural sparsity, yielding additional computational savings without explicit pruning criteria or auxiliary regularization. We further show that this behavior is not incidental but can be modulated through simple architectural choices.

Overall, our results demonstrate that tree-structured feed-forward layers constitute a practical and scalable alternative to dense FF blocks and MoE architectures.

\paragraph{Contributions.}
This work provides the first systematic investigation of hard-routed tree-structured feed-forward layers in large transformer models. Our contributions are:
(i) we demonstrate, for the first time, that FFF layers scale to billion-parameter autoregressive language models while maintaining performance comparable to dense feed-forward baselines;
(ii) we identify and analyze an emergent auto-pruning mechanism that converts dynamic tree routing into stable structural sparsity during training;
(iii) we show that routing imbalance and tree utilization can be controlled through simple architectural choices, without auxiliary losses or explicit expert-balancing mechanisms; and
(iv) we provide the first layer-level runtime benchmarks that quantify its efficiency trade-offs.



\section{Related Work}

Sparse and conditional computation has been widely studied as a means to scale neural networks efficiently. Early work on hierarchical mixtures of experts (HMEs) introduced tree-structured gating mechanisms to enable conditional execution in deep models \cite{jordan1994hierarchical}. More recent variants explore hierarchical or dynamic expert routing within transformer architectures \cite{zhao2021hierarchical, zhou2022mixtureofexpertswith}. Differentiable decision-tree layers have been proposed as neural network components that combine tree ensembles with end-to-end training \cite{hazimeh2020tree}, but these architectures are typically evaluated on vision or tabular benchmarks rather than large-scale transformer language models. In parallel, large-scale Mixture-of-Experts (MoE) models have demonstrated substantial reductions in per-token computation by activating only a small subset of experts for each input \cite{lepikhin2020gshard, fedus2022switchtransformers}. Recent work such as Mixture of a Million Experts (PEER) further pushes this paradigm by retrieving from extremely large pools of fine-grained experts, improving compute–performance trade-offs at scale \cite{he2024mixture}. While effective, these approaches typically rely on separate routing networks, expert balancing mechanisms \cite{lepikhin2020gshard, fedus2022switchtransformers, zhou2022mixtureofexpertswith}, and specialized infrastructure \cite{lepikhin2020gshard, gale2023megablocks}.

Another line of research focuses on contextual and activation-based sparsity. Methods such as Deja Vu predict at inference time which attention heads or MLP components are likely to be active/inactive for a given input, enabling conditional execution without retraining the model \cite{liu2023deja}. Related approaches, including ShadowLLM, explore predictor-based mechanisms to dynamically prune attention and feed-forward computation \cite{akhauri2024shadowllm}. These techniques are effective for inference-time acceleration but generally operate on pretrained dense models and do not introduce sparsity during training or alter the feed-forward architecture itself.

Sparsity has also been explored through pruning and mask-based techniques. Structured and unstructured pruning methods aim to remove redundant parameters either during or after training, often inspired by the lottery ticket hypothesis and related frameworks \cite{hoefler2021sparsity}. In addition, gated activation functions such as GLU and its variants (e.g., SwiGLU) have been widely adopted in transformer feed-forward blocks to improve expressivity and optimization stability \cite{shazeer2020glu}. While these approaches improve performance, they do not significantly reduce computation, as all neurons remain active for every input.

Tree-based neural architectures and differentiable decision-tree layers have also been proposed as learnable components within neural networks \cite{thebelt2022tree}. Such models combine tree structures with gradient-based optimization and have shown promising results in tabular and vision domains. Separately, transformer variants incorporating tree or syntactic structure as an inductive bias have been studied in representation learning and attention mechanisms \cite{wang2019tree}. However, these approaches are typically not designed as drop-in replacements for dense feed-forward blocks in large-scale language models.

Compared to these prior approaches, tree-based feed-forward layers occupy a distinct design point. Rather than introducing separate expert routers, inference-time predictors, or post-hoc pruning, Fast Feedforward (FFF) layers integrate routing and computation within a single tree-structured feed-forward module. This enables extreme, input-dependent sparsity during both training and inference, with predictable and structured activation patterns. Prior work on FFF layers has been limited to relatively small models and restricted settings, leaving open questions regarding scalability, training stability, and performance in modern GPT-style large language models.

This work addresses these gaps by providing a systematic study of tree-based feed-forward layers as drop-in replacements for dense MLP blocks in large transformer architectures, evaluating their scalability, efficiency, and training dynamics under realistic large-scale training regimes.

\section{Fast Feedforward Layer}
\label{sec:fff}

\paragraph{Node parameterization.}
An FFF layer \citep{Belck2023ExponentiallyFL} consists of $P$ trainable, perfect binary trees of depth $D$, which are evaluated in parallel (see Fig.~\ref{fig:overview}c).
The computation proceeds sequentially through each tree root-to-leaf.
Each node is indexed by a triple $(p,\ell,s)$, where $p\in\{1,\dots,P\}$ denotes the tree,
$\ell\in\{0,\dots,D\}$ the level, and $s\in\{0,\dots,2^\ell-1\}$ the position within the level.
Each node $(p,\ell,s)$ is associated with two sets of trainable parameters:
\begin{itemize}
  \item \textbf{Routing (linear\_in):}
  a weight vector $w^{\mathrm{in}}_{p,\ell,s}\in\mathbb{R}^{d_{\mathrm{in}}}$ and bias
  $b^{\mathrm{in}}_{p,\ell,s}\in\mathbb{R}$, producing a single scalar logit;
  \item \textbf{Output (linear\_out):}
  a weight vector $w^{\mathrm{out}}_{p,\ell,s}\in\mathbb{R}^{d_{\mathrm{out}}}$.
\end{itemize}

\paragraph{Forward computation.}
For each input sample $x\in\mathbb{R}^{d_{\mathrm{in}}}$ and each tree
$p\in\{1,\dots,P\}$, traversal starts at the root $s_0 = 0$.
For $\ell = 0,\dots,D$, exactly one node is evaluated:

\begin{equation}
z_{p,\ell}(x)
=
\left\langle x,\; w^{\mathrm{in}}_{p,\ell,s_\ell}\right\rangle
+
b^{\mathrm{in}}_{p,\ell,s_\ell}
\in\mathbb{R}.
\end{equation}
For $\ell = 0,\dots,D-1$, the next node index is updated via a hard routing decision:
\begin{equation}
s_{\ell+1}
=
2 s_\ell
+
\mathbf{1}\!\left[\mathrm{sg}\!\left(z_{p,\ell}(x)\right) \ge 0\right],
\end{equation}
where $\mathbf{1}(\cdot)$ is the indicator function and $\mathrm{sg}(\cdot)$ denotes the stop-gradient operator.
Consequently, each tree requires a fixed number of $D$ routing steps and the layer has $P \times D$ active nodes in total.
All active nodes contribute directly to the output, i.e.\ nodes on levels $0, \dots, D-1$ have a double role of routing and output generation.

\paragraph{Sparse output computation.}
The output $y = \mathrm{FFF}(x) \in \mathbb{R}^{d_{\mathrm{out}}}$ of the layer is obtained by summing the contributions of the visited nodes:
\begin{equation}
y
=
b^{\mathrm{out}}
+
\sum_{p=1}^{P}
\sum_{\ell=0}^{D}
\mathrm{GELU}\!\left(z_{p,\ell}(x)\right)
\; w^{\mathrm{out}}_{p,\ell,s_\ell}
\;.
\end{equation}

\begin{figure}
    \centering
    \includegraphics[width=\linewidth]{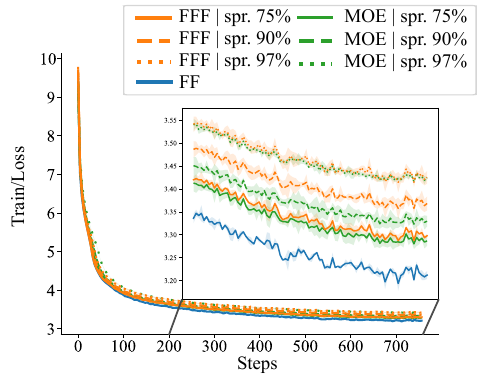}
    \caption{Training loss curves for 125M-parameter GPT-style models comparing dense feed-forward (FF), Fast Feedforward (FFF), and sparsity-matched Mixture-of-Experts (MoE) layers. }
    \label{fig:train_loss_125}
\end{figure}

\section{Autoregressive Language Modeling}
\label{sec:small_scale_comparison}

We train GPT-style language models with 125M parameters from scratch and compare dense feed-forward (FF) blocks, tree-based FFF layers, and sparsity-matched Mixture-of-Experts (MoE) layers under identical training pipelines (see Appendix~\ref{app:impl} for details). We evaluate three FFF configurations with  depth $D=3,5,7$, corresponding to approximately 75\%, 90\%, and 97\% sparsity. These models are compared against dense FF and MoEs with matched sparsity and top-$k=2$ routing.

Figure~\ref{fig:train_loss_125} shows the training loss curves for all configurations. Results are averaged over three runs with different random seeds. 
FFF achieved performance comparable to sparsity-matched MoE baselines, with only negligible differences even at high sparsity of 97\%. 
These results demonstrate that tree-based feed-forward layers can match the effectiveness of MoE at equivalent sparsity in small-scale models. Additional results in terms of perplexity (PPL) are reported in Appendix \ref{app:125_ppl}.

\section{Tree Utilization and Routing Imbalance}
\label{sec:utilization}

Expert utilization and routing balance are long-standing challenges in conditional computation models, particularly in mixture-of-experts (MoE) architectures. 
Since the variant of FFF layers used here  used no balancing mechanisms, we investigated whether similar utilization issues arise and how they manifest in practice.


\begin{figure}[ht]
    \centering
    \begin{subfigure}{0.23\textwidth}
        \centering
        \includegraphics[width=\linewidth]{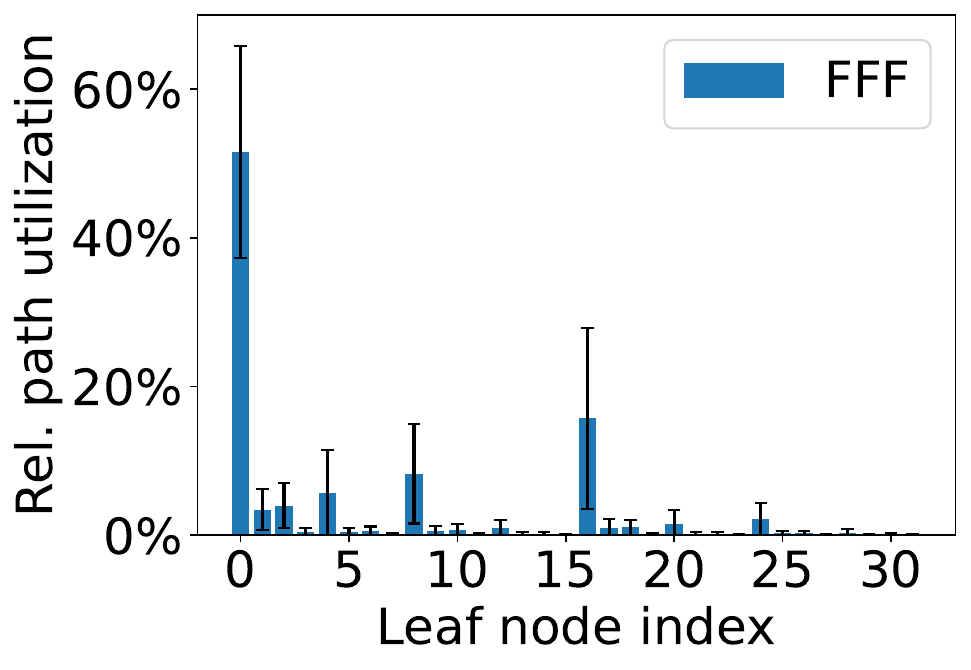}
    \end{subfigure}
    \hfill
    \begin{subfigure}{0.23\textwidth}
        \centering
        \includegraphics[width=\linewidth]{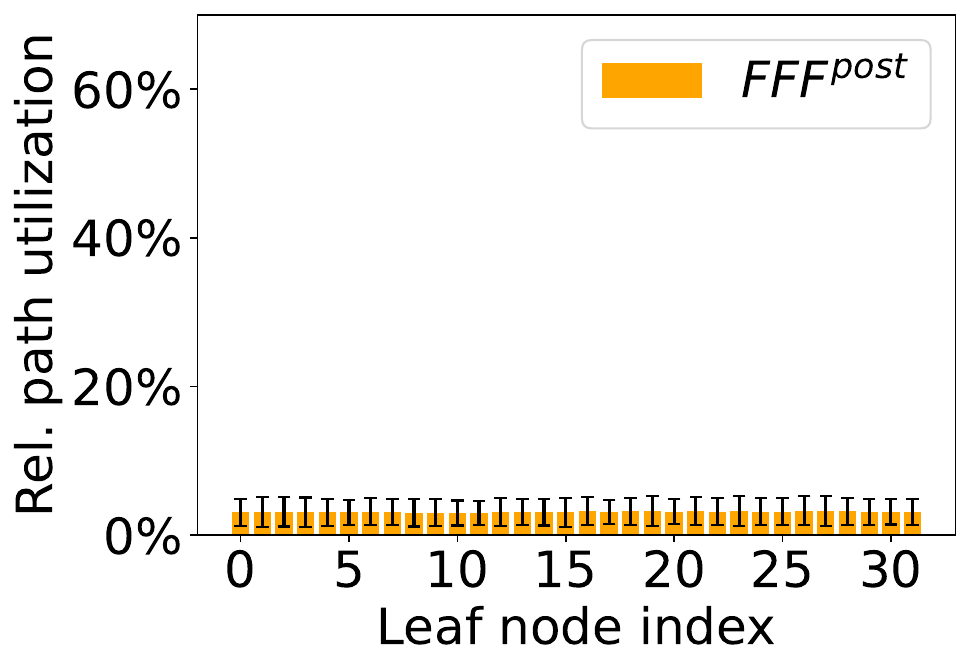}
    \end{subfigure}
    \caption{\textbf{Utilization in a d=5 tree.}
Path utilization as the normalized fraction of samples visiting each path for FFF (left) and FFF$^\text{post}$ (right). The x-axis denotes the leaf node index.}
    \label{fig:tree_util_2}
\end{figure}

We measured tree utilization by recording how frequently each path (or subtree) is visited during training and inference. Across different model sizes, tree depths, datasets, and tasks, we observe a consistent pattern of severe imbalance in tree utilization (Fig.~\ref{fig:tree_util_2}). The distribution of branch visits is highly skewed: the leftmost path from the root is selected in more than $50\%$ of cases, while the left child of the right subtree accounts for over $15\%$ of visits. The remaining paths are activated only rarely (although we found only 1\% 'dead' leaf nodes, that were never visited, for $d=3$ trees, 5\% for $d=5$ trees and 20\% for $d=7$ trees). This behavior mimics a heavy-tailed distribution over path utilization (see Appendix~\ref{app:tree-prior-analysis} for a discussion) and  mirrors expert collapse observed in MoE models trained without balancing losses.

\subsection{Architectural biases for balance / imbalance}

In MoEs imbalance typically arises from learned routers, and can be accounted for by auxiliary balancing losses. However, since FFF has no dedictaed routing structure, the observed imbalance must stem from the interaction between the tree structure, the nonlinearity, and gradient-based training of the surrounding linear layers.
In Appendix~\ref{app:autopruning} we show through a  detailed empirical and analytical investigation that this is in fact the case: the combination of hard binary routing and the GELU nonlinearity introduces an asymmetric gradient flow, where updates associated with negative node logits are strongly suppressed, while positive logits dominate the embedding space. This induces a systematic drift in node logits that biases routing decisions, to systemically prefer the left child (the child with lower index) over the right child, in every node.


Interestingly, our analysis also reveals that a simple change in architecture can fix this imbalance. This alternative design applies the nonlinearity after the sparse output accumulation:
\begin{equation*}
\mathrm{FFF}^{\mathrm{post}}(x)
=
\mathrm{GELU}\!\left(
b^{\mathrm{out}}
+
\sum_{p=1}^{P}
\sum_{\ell=0}^{D}
z_{p,\ell}(x)
\; w^{\mathrm{out}}_{p,\ell,s_\ell}
\right).
\end{equation*}
By adopting this post-GELU formulation, the gradient magnitude no longer directly depends on individual node logits.
As a result, the sign-dependent gradient suppression responsible for imbalanced routing is no longer present. As shown analytically in Appendix~\ref{app:autopruning} and empirically in Figure~\ref{fig:tree_util_2}, this modification leads to substantially more balanced tree utilization across paths. Notably, this balancing mechanism does not rely on auxiliary losses, explicit load constraints, or router regularization, while maintaining all computational advantages of FFF.

\begin{figure}
    \centering
    \includegraphics[width=\linewidth]{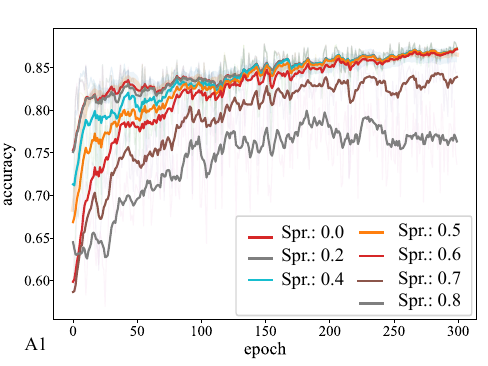}
    \caption{\textbf{Validation accuracy under statistical path pruning.}
    We evaluate different fixed pruning ratios by statistically pruning paths and measuring validation accuracy. Each curve corresponds to a different sparsity (static, permanent) ratio.}
    \label{fig:auto_acc}
\end{figure}

\subsection{Utilizing Imbalance: Auto-Pruning in Tree-Based Feed-Forward Layers}
\label{sec:auto_pruning}

When we evaluated the balanced FFF version, we found to our surprise that task performance did not depend strongly on how balanced the trees were. In fact, our results presented in Appendix~\ref{app:125_ppl} show that the balanced FFF version even slightly underperformed the imbalanced version. This suggests the the sparse training used for FFF automatically leads to a competition for resources between tree paths and a uniform distribution of resources may not be the ideal strategy. We therefore wondered whether the observed imbalance could serve a computational advantage. As optimization progresses, certain paths of the tree become effectively and permanently inactive, resulting in a form of static structural sparsity on top of the input-dependent routing mechanism.
This phenomenon manifests as a progressive concentration of computation on a subset of tree paths, while other paths receive diminishing gradient updates and eventually cease to contribute to the model output. Importantly, this auto-pruning behavior emerges automatically during training and arises from the interaction between hard tree routing decisions and the asymmetric nonlinearity of GELU, a combination that is specific to the proposed layer design.

Figure~\ref{fig:auto_acc} illustrates the effect of path pruning on validation performance. We observe that, as training proceeds, models with increasing levels of permanent path deactivation can closely match the performance of their unpruned counterparts. This indicates that auto-pruning preserves the most informative computation paths while removing redundant ones. Overall, auto-pruning provides an additional source of efficiency by converting dynamic routing decisions into stable, data-dependent structural sparsity during training. While not the primary focus of this work, this effect helps explain the robustness and scalability of tree-based feed-forward layers observed in our experiments.
In Appendix~\ref{app:autopruning} we further evaluate the auto-pruning mechanism for the OPT 125M model, showing that pruning can be an effective mechanism to further sparsity deeper FFF layers.

\begin{table}[ht]

    \caption{Perplexity (PPL) comparison of OPT-1.3B variants across two datasets and different training regimes and FFF configurations. Lower is better.
}
    \begin{center}
                \resizebox{0.48\textwidth}{!}{%
                \begin{tabular}{lccc}
                \toprule
                \textbf{Model} & \textbf{SlimPajama} & \textbf{WikiText}  \\
                \midrule
                OPT-1.3B (300B tokens) & 13.11 & 14.3  \\
                \midrule
                OPT-1.3B (26B tokens) & 15.83 & 17.2  \\
                OPT-1.3B FFF(d=4, 26B tokens) & 16.61 & 18.10  \\
                OPT-1.3B FFF(d=4, 26B tokens, FT) & \textbf{13.37} & \textbf{14.55} \\
                OPT-1.3B FFF(d=6, 26B tokens) & 17.57 & 20.01  \\
                OPT-1.3B FFF(d=6, 26B tokens, FT)& 14.11 & 15.02  \\
                \bottomrule
                \end{tabular}%
                }
    \end{center}
    \label{tab:perplexity_results}
\end{table}

\begin{table*}[ht]
\centering
\caption{Zero-shot performance on commonsense reasoning tasks. $\dagger$ Results as reported in the original papers and not reproduced in this work, i.e. OPT-1.3 (300B tokens): \cite{zhang2022opt}; Pythia: \cite{biderman2023pythia} and TinyLLaMA: \cite{zhang2024tinyllama}.}
\resizebox{\textwidth}{!}{%
\begin{tabular}{lcccccccc}
\toprule
Model & HellaSwag & Obqa & WinoGrande & ARC-c & ARC-e & boolq & piqa & Avg \\
\midrule
OPT-1.3B (300B tokens)$\dagger$ & 53.65 & 33.40 & 59.59 & 29.44 & 50.80 & 60.83 & 72.36 & 51.44 \\
\midrule
Pythia-1.0B (1T tokens)$\dagger$ & 47.16 & 31.40 & 53.43 & 27.05 & 48.99 & 57.83 & 69.21 & 48.30 \\
Pythia-1.4B (1T tokens)$\dagger$ & 52.01 & 33.20 & 57.38 & 28.50 & 54.00 & \textbf{63.27} & 70.95 & 51.33 \\
TinyLlama v1.0 (2T tokens)$\dagger$ & 59.20 & 36.00 & 59.12 & 30.12 & 55.25 & 57.83 & 73.29 & 52.99 \\
TinyLlama v1.1 (2T tokens)$\dagger$ & \textbf{61.47} & \textbf{36.80} & 59.43 & \textbf{32.68} & \textbf{55.47} & 55.99 & \textbf{73.56} & \textbf{53.63} \\
\midrule
OPT-1.3B FF & 33.7 & 27.2 & 49.6 & \textbf{24.4} & 41.7 & 59.9 & 65.2 & 43.1 \\
OPT-1.3B FFF (d=4, 26B tokens) & 32.7 & \textbf{27.2} & 50.1 & 24.1 & 40.4 & \textbf{61.5} & 65.0 & 43.0 \\
OPT-1.3B FFF (d=4, 26B tokens, FT) & \textbf{33.7} & 16.8 & \textbf{53.0} & 22.7 & \textbf{52.5} & 60.7 & \textbf{67.6} & \textbf{43.8} \\
OPT-1.3B FFF (d=6, 26B tokens) & 28.8 & 15.2 & 52.6 & 19.3 & 44.4 & 61.4 & 62.8 & 40.6 \\
OPT-1.3B FFF (d=6, 26B tokens, FT) & 33.2 & 17.4 & 52.4 & 21.7 & 50.6 & 58.7 & 67.6 & 43.0 \\
\bottomrule
\end{tabular}%
}
\label{tab:zeroshot_commonsense}
\end{table*}

\begin{table}[ht]
\centering
    \caption{Few-shot performance on reasoning and comprehension tasks. 
    Results are accuracy (\%) for MMLU (5-shot) and F1 (\%) for DROP (3-shot).$\dagger$ indicates results reported from the original papers and not reproduced in this work.}
    \resizebox{0.48\textwidth}{!}{%
    \begin{tabular}{lcc}
        \toprule
        \textbf{Model} & \textbf{MMLU } & \textbf{DROP } \\
        \midrule
        OPT-1.3B (300B tokens) & 24.90 & 14.32 \\
        \midrule
        Pythia-1.0B (1T tokens)$\dagger$ & 25.70 & 4.25 \\
        Pythia-1.4B (1T tokens)$\dagger$ & 25.41 & 12.27 \\
        TinyLlama v1.0 (2T tokens)$\dagger$ & 25.34 & 15.34 \\
        TinyLlama v1.1 (2T tokens)$\dagger$ & 26.58 & 15.31 \\
        \midrule
        OPT-1.3B FF (retrained) & 25.87 & 15.27 \\
        OPT-1.3B FFF (d=4, 26B tokens)& 25.51 & 15.25 \\
        OPT-1.3B FFF (d=4, 26B tokens, FT) & 26.4 & 15.29 \\
        OPT-1.3B FFF (d=6, 26B tokens) & 25.8 & 14.25 \\
        OPT-1.3B FFF (d=6, 26B tokens, FT) & \textbf{26.7} & 15.28 \\
        \bottomrule
    \end{tabular}%
    }
    \label{tab:fewshot_results}
\end{table}

\section{Scaling Experiments}
\label{sec:experiments}

We have now developed the necessary tools to address the core goal of this work: to evaluate the scalability and practicality of the proposed Fast Feedforward (FFF) layer as an alternative to standard feed-forward blocks in large transformer models, that maintains high levels of sparsity during inference and training. 

\subsection{Language Modeling Setup}

\paragraph{Model architecture.}
We adopt a GPT-style transformer architecture for autoregressive language modeling \cite{zhang2022opt}. In all experiments, the attention blocks remain structurally unchanged. We replace the conventional MLP layer following each attention block with the proposed FFF layer, using a hardware-compatible masked implementation 
(see Appendix~\ref{app:for-backward implementation}).

\paragraph{Training data and objective.}
Models are trained using the standard next-token prediction objective on a large-scale language modeling corpus. Given a sequence of tokens $(x_1,\dots,x_T)$, training minimizes the cross-entropy loss
\begin{equation}
\mathcal{L}
=
-\sum_{t=1}^{T} \log p(x_t \mid x_{<t}).
\end{equation}
All models were trained with a ratio of ca. $20:1$ for the number of training tokens to parameters, following \cite{hoffmann2022training}.

We consider two training regimes. In the first, models were trained entirely from scratch, with all parameters initialized randomly. In the second regime, which we refer to as \emph{fine-tuning} (FT), attention layers and other trainable components were initialized from a pretrained dense model, while the feed-forward blocks were replaced by the proposed sparse routing layers, with random initial parameters. Although replacing even a single layer requires retraining the full model, this setup reflects a practical scenario in which a pretrained model is adapted by swapping in a sparse feed-forward component as a drop-in replacement.

To contextualize our results, we report performance for several pretrained reference models trained on substantially larger corpora. In particular, we include a pretrained OPT model trained on approximately 300B tokens \cite{zhang2022opt}, as well as other publicly available models such as Pythia \cite{biderman2023pythia} and TinyLLaMA \cite{zhang2024tinyllama}, which are trained on roughly 1T and 2T tokens, respectively. These models are provided for reference only and should not be considered direct baselines, as their training budgets far exceed those used in our experiments.

Our primary baselines are dense transformer models trained from scratch using the same architecture, optimization pipeline, and number of training tokens as our sparse models. This allows for a controlled comparison that isolates the effect of replacing the feed-forward layers. We report results for two main model sizes: a 125M-parameter model evaluated on language modeling and text generation tasks (see Section \ref{sec:small_scale_comparison} and Appendix \ref{app:125_ppl}), and a 1.3B-parameter model evaluated on both language modeling and downstream benchmarks. Additional results for larger models, including 2.7B and 6.7B parameters, are provided in the appendix. Technical details about implementation, optimization, and hyperparameters can be found in Appendix~\ref{app:impl}.



\subsection{Language Modeling Results}

\paragraph{Training dynamics.}


 We evaluated FFF with depths 4 and 6, corresponding to approximately 83\% and 94\% sparsity, respectively, on a 1.3B-parameter OPT model. Depth 6 is selected based on the benchmark results in Section~\ref{sec:efficient-implementation} as a favorable trade-off between sparsity and runtime efficiency, while depth 4 provides a more conservative sparsity setting. In addition to training from scratch, we also report fine-tuning results for both configurations, where attention layers are initialized from a pretrained dense model. Across all settings, the models with FFF layer maintain stable training dynamics and closely follow the dense feed-forward baseline (see Appendix~\ref{app:training_curves_13b}), indicating that highly sparse tree-based routing layers can be trained reliably and integrated into large language models under both training regimes.

\paragraph{Evaluation on held-out data.}
For quantitative evaluation, we report perplexity on a standard held-out language modeling benchmark. Perplexity \cite{fang2024wrong} is defined as

\begin{equation}
\mathrm{PPL}
=
\exp\!\left(
\frac{1}{T}\sum_{t=1}^{T} -\log p(x_t \mid x_{<t})
\right).
\end{equation}
As shown in Table~\ref{tab:perplexity_results}, FFF models at 1.3B parameter scales achieve perplexities comparable to their dense baselines. Performance differences are typically within $10\%$, indicating that tree-based sparsification does not substantially degrade language modeling quality. 

\subsection{Zero-Shot and Few-Shot Evaluation}

To assess generalization beyond the training distribution, we evaluate models on a diverse suite of downstream language understanding tasks without task-specific fine-tuning. We report both zero-shot and few-shot accuracy across multiple benchmarks, to reduce the impact of task-specific variance or chance.

Tables~\ref{tab:zeroshot_commonsense} and \ref{tab:fewshot_results} summarizes zero-shot and few-shot performance. Across most tasks, FFF closely matches or exceeds the performance of dense feed-forward baselines when trained or fine-tuned on the same amount of data. In several cases, the tree-based variants outperform the dense model despite operating in significantly higher sparsity regimes.

We additionally include a pretrained OPT model trained on approximately 300B tokens as a reference point. Due to its substantially larger training budget compared to our models (trained on roughly 26B tokens), this model performs better on some tasks that benefit from broader topic coverage — such as Obqa. Nevertheless, on multiple benchmarks (see e.g.\ ARC-e) all FFF layer configurations, including highly sparse variants, outperform the pretrained OPT model. Similar trends are observed across other tasks, where fine-tuned FFF models consistently achieve equal or higher accuracy than their dense counterparts.

Notably, results on MMLU (Table \ref{tab:fewshot_results}), reported as an average across all subtasks, show that sparser FFF models can achieve higher accuracy than dense feed-forward baselines. These findings indicate that tree-based sparse routing not only scales effectively to large language models, but can also preserve, and in some cases improve, generalization and in-context learning performance, even under extreme sparsity and reduced training data.



\paragraph{Beyond training from scratch and fine-tuning in language models.} We also evaluated the proposed sparse routing layer in additional settings to further assess its applicability. In particular, we applied FFF to a transformer-based object detection model (DETR), showing that the sparse layers match the performance of their dense counterparts and exhibit stable training behavior. Detailed results are reported in the Appendix~\ref{app:vision} In Appendix~\ref{app:large-scale}, we present additional results for 2.7B and 6.7B parameter OPT models showing promising results for the $d=4$ FFF variant.



\section{Efficiency of Tree-Based Feed-Forward Layers}
\label{sec:efficient-implementation}

Efficiency in terms of both runtime and computational cost is a central motivation for introducing tree-based feed-forward layers in large-scale models. While the Fast Feedforward (FFF) layer enables high structured sparsity, it also introduces a sequential, depth-wise execution pattern that differs fundamentally from the fully parallel execution of dense feed-forward blocks. This raises an important practical question: can such a highly sparse but hierarchical layer be computationally competitive, or even advantageous, at scale?

To address this question, we benchmark the execution of FFF under its intended sparse computation model and compare it against a dense feed-forward baseline. Rather than focusing on kernel-level optimizations on current accelerators, our goal is to evaluate whether the substantial reduction in computation enabled by tree-based sparsity can outweigh the overhead of sequential traversal.

\begin{table*}[t]
\centering
\caption{Efficiency metrics for three selected model configurations, including MLP-layer sparsity, overall model sparsity, relative FLOPs, and relative FLOPs/S compared to a dense GPT baseline.}
\label{tab:sparsity_stats}
\begin{tabular}{lcccc}
\toprule
Model & Overall Sparsity & MLP block Sparsity & Relative FLOPs & Relative FLOPs/S \\
\midrule
Dense FF (GPT) & 0\% & 0\% & 1.0$\times$ & 1.0$\times$ \\
FFF (d=13) & 61\% & 99\% & 0.0016$\times$ & 0.0084$\times$ \\
FFF (d=6) & 58\% & 94\% & 0.0952$\times$ & 0.45$\times$ \\

FFF (d=4) & 52\% & 83\% & 0.2666$\times$ & 0.78$\times$ \\
\bottomrule
\end{tabular}
\end{table*}


\begin{figure}[ht]
    \centering
    \includegraphics[width=\linewidth]{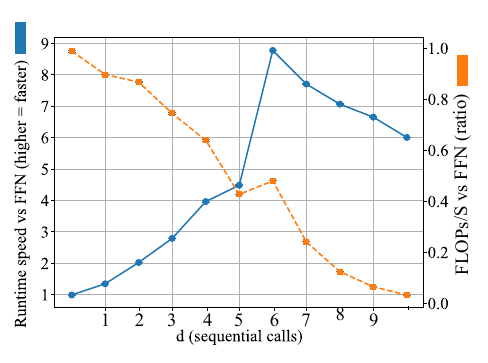}
    \caption{Relative speedup and FLOPS/s for all configurations, ranging from depth 0 (dense FF) to depth 11. Results are averaged over five runs. d indicate depth of tree.}
    \label{fig:speedup simulation}
\end{figure}



\begin{table}[t]
\centering
\caption{Speedup relative to the dense GPT baseline for three selected model on single A100 configurations.}
\label{tab:speedup}
\begin{tabular}{lcc}
\toprule
Model & layer execution (ms)& Speedup \\
\midrule
Dense FF (GPT) & 3.03 & 1.0$\times$ \\
FFF (d=13) & 0.57 & 5.3$\times$ \\
FFF (d=6) & 0.35 & 8.7$\times$ \\
FFF (d=4) & 1.08& 2.8$\times$ \\
\bottomrule
\end{tabular}
\end{table}

\subsection{Runtime Speedup}

We evaluate the runtime behavior of tree-based feed-forward layers under their sequential sparse execution model, in which only the nodes along the active traversal path are evaluated and all inactive paths incur no computation. Runtime is measured at the layer level and compared against a dense feed-forward baseline under identical conditions.


As shown in Fig.~\ref{fig:speedup simulation} and summarized in Table~\ref{tab:speedup}, increasing tree depth leads to substantial runtime improvements despite the introduction of depth-wise sequential traversal. For configurations relevant to large language models, a tree depth of 6 achieves the highest speedup, with an average improvement of approximately $8.7\times$ over dense feed-forward layers. While deeper trees further reduce computation, diminishing returns and increased traversal overhead limit additional gains.

These results identify depth 6 as a practical trade-off between extreme sparsity and efficient execution. We therefore adopt this configuration in most of our large-scale experiments, alongside shallower settings that prioritize reduced routing depth.


\subsection{Sparsity and FLOPs Analysis}

We characterize sparsity and computational cost at both the layer and model levels. \emph{Feed-forward layer sparsity} measures the fraction of inactive nodes within the tree-based feed-forward layers and isolates the sparsity induced by the proposed architecture. \emph{Effective model sparsity} captures the resulting reduction in computation across the full transformer, accounting for the fact that attention layers remain dense.

As summarized in Table~\ref{tab:sparsity_stats}, sparsity grows exponentially with tree depth. Even shallow configurations achieve high sparsity (e.g., approximately 83\% for $d=4$), while deeper trees reach extreme regimes, exceeding 94\% sparsity at depth 6 and more than 99\% at depth 13. This behavior follows directly from the hierarchical structure of the trees and the activation of a single path per input.

To complement the runtime measurements, we also report analytical estimates of per-token floating-point operations (FLOPs), computed based on the number of activated nodes along each routing path. Table~\ref{tab:sparsity_stats} reports FLOPs normalized to a dense feed-forward baseline and confirms an exponential reduction with increasing tree depth, consistent with the observed sparsity trends.

While total FLOPs decrease sharply with increasing sparsity, measured throughput in terms of FLOPs per second also decreases. This behavior is expected and desirable: as the amount of computation performed per token is reduced, the aggregate FLOPs executed per second naturally drop. In this regime, FLOPs/s reflects the reduced computational demand rather than a slowdown of the layer itself, as demonstrated by the measured speedups, where wall-clock runtime improves despite the lower FLOPs/s.




\section{Discussion}




In this work, we studied tree-based sparse feed-forward layers 
as structured alternatives to dense feed-forward (FF) layers, demonstrating that they perform on par with mixture-of-experts models. This indicates that tree-based routing can serve as a practical, scalable, and predictable sparsification mechanism. We demonstrated that replacing dense FF layers with tree-based sparse layers scales effectively to large language models and generalizes across tasks and domains while maintaining performance close to dense baselines.

Across multiple model sizes and evaluation settings, we showed that these layers can operate in high sparsity regimes—often exceeding 90\%—while achieving comparable and, in several cases, superior performance relative to dense FF baselines trained under the same data and optimization budgets. Notably, we observed that sparse tree-based models can outperform dense counterparts despite activating only a small fraction of parameters, highlighting the effectiveness of structured, input-dependent computation.

Beyond empirical performance, we analyzed the routing dynamics induced by hierarchical binary trees and identified an emergent \emph{auto-pruning} behavior. This mechanism converts dynamic, input-dependent routing into stable structural sparsity, enabling permanent deactivation of redundant computation paths without auxiliary losses or explicit pruning criteria. We further showed that routing imbalance—often viewed as a limitation in sparse architectures—is not necessarily harmful in tree-based routing and can be controlled through simple architectural choices when needed.

While our results indicate strong potential for both scalability and efficiency, we note that the practical runtime benefits of such layers are currently constrained by hardware and software support. Many highly optimized components in large language models rely on accelerator-specific kernels, whereas dynamic, tree-based sparse execution remains underexplored in current hardware stacks. As such, our efficiency results are intended to highlight theoretical and simulated potential rather than immediate deployment-ready speedups.

Overall, this work aims to encourage further exploration of tree-based sparse computation as a viable direction for efficient large-scale models. We believe that the combination of dynamic, predictable, and structured sparsity, along with emerging auto-pruning properties, makes tree-based feed-forward layers a promising foundation for future research in scalable and efficient neural architectures.

\section*{Acknowledgements}
This work was funded by the German Federal Ministry for Economic Affairs and Climate Action (BMFTR) project ESCADE (01MN23004D). DK is funded by project SAIL (grant no. NW21-059A). The authors gratefully acknowledge the Gauss Centre for Supercomputing e.V. (\url{www.gauss-centre.eu}) for funding this project by providing computing time on the GCS Supercomputer JUWELS at Jülich Supercomputing Centre (JSC).



\section*{Impact Statement}

This work aims to advance the field of machine learning by studying structured sparse feed-forward layers for modern transformer architectures. From an ethical perspective, the proposed tree-based design introduces a more structured and interpretable form of conditional computation, which may help improve understanding and transparency of model behavior compared to fully dense architectures.

From a societal standpoint, large transformer models are computationally and energy intensive. By enabling extreme and dynamic sparsity while preserving performance, this work contributes toward more computationally efficient model designs, which may help reduce energy consumption and environmental impact during training and deployment at scale. While the work does not directly address downstream applications or deployment risks, it encourages future research on efficient, scalable, and responsible large-scale machine learning systems.

\nocite{wang2024mmlu, dua2019drop, zellers2019hellaswag, myrzakhan2024open, sakaguchi2021winogrande, clark2018think, clark2019boolq, bisk2020piqa} 

\bibliography{example_paper}
\bibliographystyle{icml2026}




\newpage
\appendix
\onecolumn
\section{Additional Perplexity Results}
\label{app:125_ppl}
This appendix provides additional evaluation results. Table~\ref{tab:ppl_in_by_group} reports perplexity (PPL) values for various OPT-125M parameter model configurations trained from scratch.


\begin{table}[h]
\centering
\caption{Per-run scalar $\mathrm{PPL}$ aggregated by strategy and subgroup. Entries are mean $\pm$ std, averaged over multiple runs with different random seeds (best). Lower is better.}
\label{tab:ppl_in_by_group}
\begin{tabular}{llc}
\toprule
\textbf{Variant} & \textbf{Tree depth / \#experts} & \textbf{$\mathrm{PPL}$ (best)} \\
\midrule
FF  & --  & $23.7786 \pm 0.0493$ (23.7409) \\
\midrule
FFF & 3   & $25.9104 \pm 0.0690$ (25.8447) \\
    & 5   & $27.6623 \pm 0.3203$ (27.2926) \\
    & 7   & $29.3686 \pm 0.2763$ (29.0518) \\
\midrule
FFF$^{\text{post}}$ & 3 & $29.6824 \pm 0.0203$ (29.6640) \\
                    & 5 & $29.8080 \pm 0.5093$ (29.0890) \\
                    & 7 & $30.1630 \pm 0.5413$ (29.6201) \\
\midrule
MoE & 8   & $25.8318 \pm 0.3065$ (25.4780) \\
    & 21  & $26.7035 \pm 0.3507$ (26.3004) \\
    & 106 & $29.3075 \pm 0.1130$ (29.2016) \\
\bottomrule
\end{tabular}
\end{table}
\section{Training Dynamics for the 1.3B Model}
\label{app:training_curves_13b}

This appendix presents training loss curves for the 1.3B-parameter language model to complement the evaluation results reported in the main text. While quantitative metrics for this model size are discussed in Section~\ref{sec:experiments}, the plots here provide additional insight into optimization stability and training behavior.

\begin{figure}[h]
    \centering
    \includegraphics[width=\linewidth]{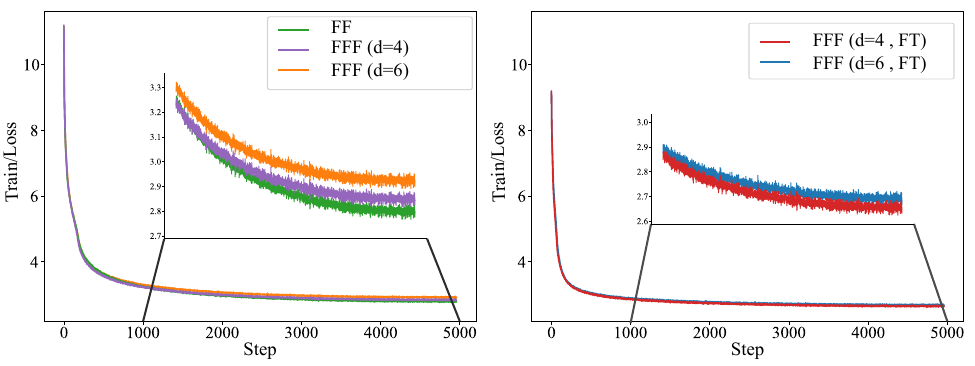}
    \caption{training curves 1.3 billion parameter}
    \label{fig:train_loss_1B}
\end{figure}

Figure~\ref{fig:train_loss_1B} shows two training regimes. The left panel reports training from scratch for three configurations: a dense feed-forward (FF) baseline and two Fast Feedforward (FFF) variants with tree depths 4 and 6. All models are trained using the same pipeline and budget. The curves demonstrate that both FFF configurations closely track the dense baseline throughout training, indicating stable optimization even at high sparsity levels.

The right panel shows the fine-tuning regime, where the dense FF layers of a pretrained OPT-1.3B model are replaced with FFF layers, while all remaining weights are initialized from the pretrained checkpoint. Training remains stable, and the loss rapidly converges to the expected range, demonstrating that FFF layers can be integrated into pretrained large language models and adapted efficiently through fine-tuning.

\section{Additional Large-Scale Training Results}
\label{app:large-scale}


\begin{figure}[h]
    \centering
    \includegraphics[width=\linewidth]{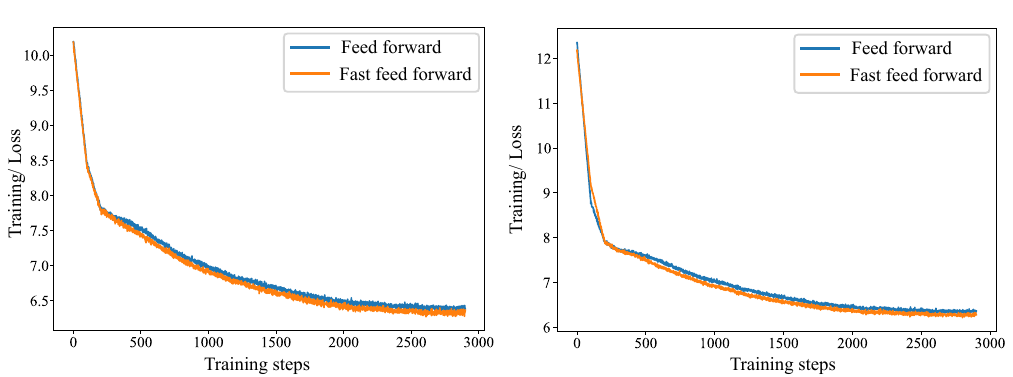}
    \caption{Training loss curves for OPT-style models with 2.7B and 6.7B parameters trained from scratch on a reduced SlimPajama subset. For each model size, we compare a dense feed-forward baseline with a Forest Router configuration of depth 4 (approximately 84\% sparsity). Due to compute constraints, these experiments are conducted on a limited dataset and are intended to evaluate scalability and training stability rather than final convergence.}
    \label{fig:opt_large_training}
\end{figure}

To further evaluate the scalability of tree-based feed-forward layers, we conduct additional training experiments on larger OPT-style models with 2.7B and 6.7B parameters. Due to computational constraints, these models are trained from scratch in a controlled environment using a reduced subset of the SlimPajama dataset \cite{cerebras2023slimpajama}. As a result, these experiments are not intended to provide a direct comparison with the models trained on the full dataset, but rather to assess training stability and scaling behavior.

For each model size, we compare a dense feed-forward baseline with a Forest Router configuration of depth 4, corresponding to approximately 84\% sparsity. Both models are trained using the same architecture, optimization pipeline, and data subset. Each plot reports two runs, one for the dense feed-forward model and one for FFF.

As shown in Fig.~\ref{fig:opt_large_training}, FFF models closely follow the training dynamics of dense feed-forward baselines. In both model sizes, the sparse models exhibit comparable or slightly improved loss trajectories during the early and middle stages of training. While we cannot verify whether this advantage would persist through full training on a substantially larger dataset, similar behavior has been observed in smaller models, where dense feed-forward layers ultimately achieve marginally lower final loss. Nevertheless, the small performance gap and stable optimization behavior indicate that FFF layers remain a viable replacement at these scales.

Overall, these results provide additional evidence that tree-based feed-forward layers can be scaled to multi-billion-parameter models while maintaining stable training dynamics and competitive performance.

\section{Extension to Vision Tasks}
\label{app:vision}

To evaluate whether tree-based feed-forward layers generalize beyond language modeling, we apply the same replacement strategy to a transformer-based object detection model. Specifically, we replace the feed-forward layers in a DETR architecture \cite{carion2020end} with Fast Feedforward (FFF) layers, while keeping all other components unchanged.

We train both the dense FF baseline and the FFF-based DETR model under a fixed compute budget, using the CPPE-5 dataset \cite{dagli2023cppe} for object detection. Training is performed for a fixed number of epochs due to computational constraints. All experiments are repeated over multiple runs with different random seeds, and results are reported as averages to reduce variance.

\begin{figure}[h]
    \centering
    \includegraphics[width=\linewidth]{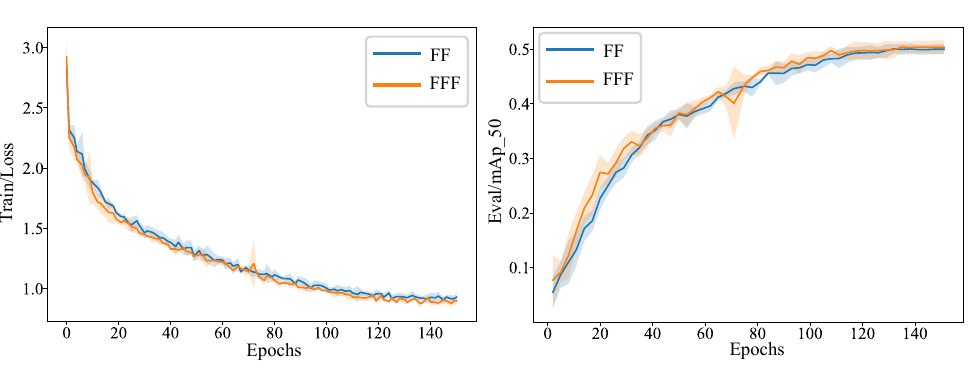}
    \caption{Extension of Fast Feedforward (FFF) layers to object detection with DETR on the CPPE-5 dataset. 
    Right: evaluation performance measured by $\mathrm{mAP}@50$, defined as mean Average Precision at an IoU threshold of 0.5. 
    Left: training loss curves. Results are averaged over multiple runs with different random seeds. FFF-based models achieve performance comparable to dense feed-forward baselines under a fixed training budget.}
    \label{fig:detr_results}
\end{figure}

Figure~\ref{fig:detr_results} summarizes the results. The left panel shows the training loss curves, while the right panel reports evaluation performance measured by $\mathrm{mAP}@50$, i.e., mean Average Precision computed at an intersection-over-union (IoU) threshold of 0.5, a standard metric for object detection. Across runs, the FFF-based model closely matches the performance of the dense feed-forward baseline, and in some cases slightly outperforms it.

Due to limited compute, we do not compare against long-horizon training or fully pretrained DETR models. Nevertheless, these results indicate that FFF layers can serve as effective drop-in replacements for dense feed-forward blocks in vision transformers, extending the applicability of tree-based sparse computation beyond language models.

\section{CDF--PDF relation and monotonicity of $p=\mathbb{P}(z>0)$}
\label{app:cdf_pdf_monotone}

Let $z\sim\mathcal{N}(\mu,\sigma^2)$ with $\sigma>0$ and define $p=\mathbb{P}(z>0)$.
Standardize $\tilde z=(z-\mu)/\sigma\sim\mathcal{N}(0,1)$:
\begin{equation}
p=\mathbb{P}(z>0)
=\mathbb{P}\!\left(\tilde z>\frac{-\mu}{\sigma}\right)
=1-\Phi\!\left(\frac{-\mu}{\sigma}\right)
=\Phi\!\left(\frac{\mu}{\sigma}\right),
\end{equation}
where $\Phi$ is the CDF of the standard normal distribution.

\paragraph{CDF and PDF.}
For any continuous random variable, the CDF is the integral of its PDF:
\begin{equation}
\Phi(x)=\int_{-\infty}^{x}\phi(t)\,dt,
\qquad\Rightarrow\qquad
\frac{d}{dx}\Phi(x)=\phi(x).
\end{equation}
For the standard normal,
\begin{equation}
\phi(x)=\frac{1}{\sqrt{2\pi}}\exp\!\left(-\frac{x^2}{2}\right)>0
\quad \text{for all } x\in\mathbb{R}.
\end{equation}

\paragraph{Monotonicity in the mean.}
Using the chain rule,
\begin{equation}
\frac{\partial p}{\partial \mu}
=\frac{\partial}{\partial \mu}\Phi\!\left(\frac{\mu}{\sigma}\right)
=\phi\!\left(\frac{\mu}{\sigma}\right)\cdot \frac{1}{\sigma}.
\end{equation}
Since $\sigma>0$ and $\phi(\cdot)>0$, we obtain $\frac{\partial p}{\partial \mu}>0$, hence
$p$ increases monotonically with $\mu$.

\section{FFF Block: Forward and Backward}

\label{app:for-backward implementation}

To simplify the theoretical analysis and implementation for the large-scale experiments, we use a masked implementation of the FFF layer that is fully compatible with existing GPU kernels. This formulation is functionally equivalent to the sequential tree traversal described in the main text: it produces identical forward activations and backward gradients for all parameters.

\paragraph{FFF layer.}
Let $X \in \mathbb{R}^{B \times d_{\mathrm{in}}}$ with $B$ and $d_{\mathrm{in}}$ representing the batch size and input dimensionality, respectively.
We project to per-tree node logits
\begin{equation}
Z = \mathrm{reshape}(XW_{\mathrm{in}} + \mathbf{b}_{\mathrm{in}}) \in \mathbb{R}^{B \times P \times N},
\end{equation}
where $P$ and $N$ refer to the number of parallel trees and nodes per tree, respectively.
We define a stop-gradient operator $\mathrm{sg}(\cdot)$ such that
$\frac{\partial\,\mathrm{sg}(u)}{\partial u}=0$ and $\mathrm{sg}(u)=u$ in the forward pass.
The tree routing process, i.e.\ traversing the tree level by level and choosing a node based on the previous layer's node output, is represented by the binary-tree search (BTS) operator. It produces a binary mask
\begin{equation}
M = \mathrm{BTS}(\mathrm{sg}(Z)) \in \{0,1\}^{B \times P \times N}, 
\end{equation}
that can be multiplied element-wise with $Z$ to select the outputs from the active nodes. Note that $Z$ is computed directly from the input for all nodes and afterwards BTS realizes the tree structure by selecting an active path down the tree. Conceptually, BTS can be considered a specific form of expert selection module that shares its parameters with the experts represented by individual tree paths. This view on FFF is functionally identical to the formulation in the main text, but greatly simplifies the theoretical analysis presented here. Additionally, experts share some of their parameters where ever the corresponding paths overlap.

To produce the FFF block's output, we apply the nonlinearity and then mask:
\begin{equation}
G=\mathrm{GELU}(Z),
\qquad
A = M \odot G.
\end{equation}
Flattening across $(P,N)$ yields $\bar{A}=\mathrm{flatten}(A)\in\mathbb{R}^{B\times(PN)}$, and the layer output is
\begin{equation}
Y = \bar{A}W_{\mathrm{out}} + \mathbf{b}_{\mathrm{out}}.
\end{equation}
Equivalently with omitting the reshape operations for simplicity,
\begin{equation}
\mathrm{FFF}(X)=\mathrm{linear\_out}\!\left(\mathrm{BTS}(\mathrm{sg}(Z)) \odot \mathrm{GELU}(Z)\right),
\quad Z=\mathrm{linear\_in}(X).
\end{equation}

\paragraph{Backward pass (no gradients through BTS).}
Let $\triangledown_Y=\frac{\partial \mathcal{L}}{\partial Y}$.
Backprop through the output projection gives
\begin{equation}
\triangledown
_{\bar{A}} = \triangledown_Y W_{\mathrm{out}}^\top,
\quad
\frac{\partial \mathcal{L}}{\partial W_{\mathrm{out}}}=\bar{A}^\top \triangledown_Y,
\quad
\frac{\partial \mathcal{L}}{\partial \mathbf{b}_{\mathrm{out}}}=\sum_{b=1}^B \triangledown_{Y,b}.
\end{equation}
Unflatten $\triangledown_{\bar{A}}$ to $\triangledown_A\in\mathbb{R}^{B\times P\times N}$.
Since $A=M\odot G$ and $M=\mathrm{BTS}(\mathrm{sg}(Z))$, we enforce $\frac{\partial M}{\partial Z}=0$, hence
\begin{equation}
\triangledown_G \equiv \frac{\partial \mathcal{L}}{\partial G} = M \odot \triangledown_A.
\end{equation}
With $G=\mathrm{GELU}(Z)$, we obtain
\begin{equation}
\triangledown_Z \equiv \frac{\partial \mathcal{L}}{\partial Z}
= \triangledown_G \odot \mathrm{GELU}'(Z)
= (M \odot \triangledown_A)\odot \mathrm{GELU}'(Z).
\end{equation}
Flatten $\triangledown_Z$ to $\triangledown_{\mathrm{flat}}=\mathrm{flatten}(\triangledown_Z)\in\mathbb{R}^{B\times(PN)}$. Then
\begin{equation}
\frac{\partial \mathcal{L}}{\partial W_{\mathrm{in}}}=X^\top \triangledown_{\mathrm{flat}},
\quad
\frac{\partial \mathcal{L}}{\partial \mathbf{b}_{\mathrm{in}}}=\sum_{b=1}^B \triangledown_{\mathrm{flat},b},
\quad
\frac{\partial \mathcal{L}}{\partial X}=\triangledown_{\mathrm{flat}} W_{\mathrm{in}}^\top.
\end{equation}

\paragraph{FFF$^{\mathrm{\textbf{post}}}$ layer.}
Let $X \in \mathbb{R}^{B \times d_{\mathrm{in}}}$ and
\begin{equation}
Z = \mathrm{reshape}(XW_{\mathrm{in}} + \mathbf{b}_{\mathrm{in}}) \in \mathbb{R}^{B \times P \times N}.
\end{equation}
We use a stop-gradient operator $\mathrm{sg}(\cdot)$ with
$\frac{\partial\,\mathrm{sg}(u)}{\partial u}=0$.
The binary-tree search produces a hard mask
\begin{equation}
M = \mathrm{BTS}(\mathrm{sg}(Z)) \in \{0,1\}^{B \times P \times N}.
\end{equation}
We apply masking in the projected space,
\begin{equation}
A = M \odot Z,
\end{equation}
then project back and apply the nonlinearity:
\begin{equation}
U=\mathrm{linear\_out}(A), \qquad Y=\mathrm{GELU}(U).
\end{equation}
Equivalently,
\begin{equation}
\mathrm{FFF}^{\mathrm{post}}(X)
=
\mathrm{GELU}\!\left(\mathrm{linear\_out}\!\left(\mathrm{BTS}(\mathrm{sg}(Z))\odot Z\right)\right),
\quad Z=\mathrm{linear\_in}(X).
\end{equation}

\paragraph{Backward pass (no gradients through BTS).}
Let $G_Y=\frac{\partial \mathcal{L}}{\partial Y}$.
Backprop through the final GELU gives
\begin{equation}
\triangledown_U \equiv \frac{\partial \mathcal{L}}{\partial U}
=
G_Y \odot \mathrm{GELU}'(U).
\end{equation}
Backprop through the output projection yields
\begin{equation}
\triangledown_{\bar{A}} = \triangledown_U W_{\mathrm{out}}^\top,
\quad
\frac{\partial \mathcal{L}}{\partial W_{\mathrm{out}}}=\bar{A}^\top \triangledown_U,
\quad
\frac{\partial \mathcal{L}}{\partial \mathbf{b}_{\mathrm{out}}}=\sum_{b=1}^B (\triangledown_U)_b.
\end{equation}
Unflatten $\triangledown_{\bar{A}}$ to $\triangledown_A\in\mathbb{R}^{B\times P\times N}$.
Since $A=M\odot Z$ and $M=\mathrm{BTS}(\mathrm{sg}(Z))$, we enforce $\frac{\partial M}{\partial Z}=0$, hence
\begin{equation}
\triangledown_Z \equiv \frac{\partial \mathcal{L}}{\partial Z} = M \odot \triangledown_A.
\end{equation}
Flatten $\triangledown_Z$ to $\triangledown_{\mathrm{flat}}=\mathrm{flatten}(\triangledown_Z)\in\mathbb{R}^{B\times(PN)}$. Then
\begin{equation}
\frac{\partial \mathcal{L}}{\partial W_{\mathrm{in}}}=X^\top \triangledown_{\mathrm{flat}},
\quad
\frac{\partial \mathcal{L}}{\partial \mathbf{b}_{\mathrm{in}}}=\sum_{b=1}^B \triangledown_{\mathrm{flat},b},
\quad
\frac{\partial \mathcal{L}}{\partial X}=\triangledown_{\mathrm{flat}} W_{\mathrm{in}}^\top.
\end{equation}


\paragraph{Positive-branch probability under Gaussian logits.}
Assuming the node logit follows $z\sim\mathcal{N}(\mu,\sigma^2)$, the probability of selecting the positive branch is
\begin{equation}
p \triangleq \mathbb{P}(z>0)=\Phi\!\left(\frac{\mu}{\sigma}\right),
\end{equation}
where $\Phi$ is the standard normal CDF. Consequently $p$ is monotonically increasing in the mean $\mu$ (see Appendix~\ref{app:cdf_pdf_monotone}).

\paragraph{Mean-logit drift under masked GELU gating.}
Consider one node logit $z=w^\top h + b$ produced inside $\mathrm{linear\_in}$.
Define the mean logit $\mu\triangleq\mathbb{E}[z]=w^\top m+b$ where $m=\mathbb{E}[h]$.
Under SGD,
\begin{equation}
w^{+}=w-\eta \frac{\partial \mathcal{L}}{\partial w},\qquad
b^{+}=b-\eta \frac{\partial \mathcal{L}}{\partial b},
\end{equation}
hence
\begin{equation}
\mu^{+}-\mu
=(w^{+}-w)^\top m+(b^{+}-b)
=-\eta\left(m^\top \frac{\partial \mathcal{L}}{\partial w}+\frac{\partial \mathcal{L}}{\partial b}\right).
\end{equation}
Using $\frac{\partial \mathcal{L}}{\partial w}=\frac{\partial \mathcal{L}}{\partial z}\,h$ and
$\frac{\partial \mathcal{L}}{\partial b}=\frac{\partial \mathcal{L}}{\partial z}$ gives
\begin{equation}
\mathbb{E}[\mu^{+}-\mu]
=-\eta\,\mathbb{E}\!\left[\left(\frac{\partial \mathcal{L}}{\partial z}\right)(m^\top h+1)\right].
\end{equation}

For the FFF variant $Y=\mathrm{linear\_out}(M\odot \mathrm{GELU}(Z))$, we treat $M=\mathrm{BTS}(\mathrm{sg}(Z))$
as constant in backprop, i.e.\ $\partial M/\partial Z=0$.
Let $g\triangleq \partial \mathcal{L}/\partial (M\,\mathrm{GELU}(z))$ be the upstream gradient at this node.
Then
\begin{equation}
\frac{\partial \mathcal{L}}{\partial z}=g\cdot M\cdot \mathrm{GELU}'(z),
\end{equation}
and therefore
\begin{equation}
\mathbb{E}[\mu^{+}-\mu]
=-\eta\,\mathbb{E}\!\left[g\,M\,\mathrm{GELU}'(z)\,(m^\top h+1)\right].
\end{equation}
Splitting over $z>0$ and $z<0$ yields two contributions. Since $\mathrm{GELU}'(z)$ is close to $1$ for $z>0$
and substantially smaller for $z<0$, the negative-side term is typically suppressed, giving the approximation
\begin{equation}
\mathbb{E}[\mu^{+}-\mu]\approx
-\eta\,\mathbb{E}\!\left[g\,M\,(m^\top h+1)\mid z>0\right]\mathbb{P}(z>0).
\end{equation}
Empirically we observe the upstream signal $g$ has negative mean during training (Appendix~\ref{app:autopruning}),
which makes the right-hand side positive (under mild regularity/weak-correlation assumptions), implying
$\mathbb{E}[\mu^{+}-\mu]>0$. Since $p=\mathbb{P}(z>0)=\Phi(\mu/\sigma)$ and $\partial p/\partial \mu>0$
(Appendix~\ref{app:cdf_pdf_monotone}), this corresponds to $\mathbb{E}[\Delta p]>0$.

\section{Auto-Pruning Analysis}
\label{app:autopruning}

This appendix provides a detailed empirical and analytical investigation of the auto-pruning behavior observed in tree-based feed-forward layers. For simplicity, the analysis is presented using the masked formulation of the FFF layer, which is functionally equivalent to the sequential tree traversal described in the main text. We show how stable structural sparsity emerges from the interaction between hard tree routing decisions and the asymmetric nonlinearity of GELU.

\subsection{Empirical Observations}

To isolate the underlying dynamics of auto-pruning, we conduct controlled toy experiments using a large batch size of 10{,}000 to minimize stochastic effects during training.

\begin{figure}[h]
    \centering
    \includegraphics[width=\linewidth]{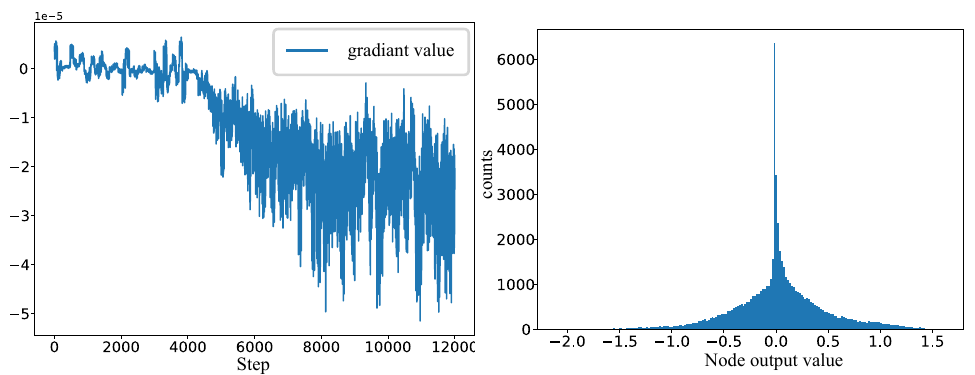}
    \caption{Empirical analysis of auto-pruning dynamics in a toy setting. 
    Left: The gradient signal value received by a randomly selected tree node during backpropagation during training, showing persistent negative values after an initial training phase. 
    Right: Histogram of the corresponding node output activations, which follow an approximately zero-mean normal distribution. 
    Both measurements are computed using batch size of 10{,}000 to reduce stochastic variability.}
    \label{fig:autopruning_analysis}
\end{figure}

The first observation concerns the distribution of activations at a randomly selected internal tree node. As shown in Fig.~\ref{fig:autopruning_analysis} (right), the node logit follows an approximately zero-mean normal distribution after training. This symmetry indicates that routing decisions are not biased toward either branch at initialization and that specialization arises through training dynamics.

The second observation analyzes the gradient signal received by the same node during backpropagation. Fig.~\ref{fig:autopruning_analysis} (left) shows the distribution of the upstream gradient after an initial transient phase. We observe that the gradient concentrates almost entirely in the negative regime. This persistent negative signal encourages the routing gate to suppress one branch consistently, driving the node toward permanent deactivation due to the combination of GELU as an asymmetric nonlinearity and the hard thresholding of the tree.

Together, these observations provide empirical evidence that motivates the analysis of the auto-pruning dynamics and mechanisms presented in the next section.

\subsection{Analytical Explanation}

Consider the routing logit for a single tree node,
\begin{equation}
z = w^\top h + b,
\end{equation}
where $h$ denotes the node input and $w,b$ are the parameters of the routing projection. We define the mean logit as
\begin{equation}
\mu \triangleq \mathbb{E}[z].
\end{equation}
Empirically, we observe that $z$ remains approximately Gaussian across training,
\begin{equation}
z \sim \mathcal{N}(\mu, \sigma^2),
\end{equation}
which motivates the following analysis.

The probability of selecting the positive branch is
\begin{equation}
p = \mathbb{P}(z > 0) = \Phi\!\left(\frac{\mu}{\sigma}\right),
\end{equation}
where $\Phi$ denotes the standard normal cumulative distribution function. Since $\Phi$ is monotonic, any positive drift in $\mu$ increases the likelihood of selecting the same branch.

During training with stochastic gradient descent, the expected change in the mean logit satisfies
\begin{equation}
\mathbb{E}[\Delta \mu]
=
-\eta\,\mathbb{E}\!\left[
\frac{\partial \mathcal{L}}{\partial z}(m^\top h + 1)
\right],
\quad
m \triangleq \mathbb{E}[h],
\end{equation}
where $\eta$ is the learning rate and $\mathcal{L}$ the task loss.

For the standard FFF formulation,
\begin{equation}
Y = \mathrm{linear\_out}(M \odot \mathrm{GELU}(Z)),
\end{equation}
the routing mask $M=\mathrm{BTS}(\mathrm{sg}(Z))$ is treated as constant during backpropagation. Let
\begin{equation}
g \triangleq \frac{\partial \mathcal{L}}{\partial (M\,\mathrm{GELU}(z))}
\end{equation}
denote the upstream gradient. The gradient with respect to $z$ becomes
\begin{equation}
\frac{\partial \mathcal{L}}{\partial z}
=
g \cdot M \cdot \mathrm{GELU}'(z).
\end{equation}

Since $\mathrm{GELU}'(z)$ is close to $1$ for $z>0$ and significantly smaller for $z<0$, gradient contributions from the negative region are suppressed. Empirically, we observe that the upstream gradient $g$ has a negative mean, yielding
\begin{equation}
\mathbb{E}[\Delta \mu] > 0,
\end{equation}
which causes $\mu$ to drift toward positive values. This reinforces the selection of the same branch and progressively suppresses alternatives.

\subsection{From Dynamic Routing to Static Sparsity}

As training progresses, nodes whose logits consistently drift negative are visited increasingly rarely. These nodes receive diminishing gradient updates and eventually become permanently inactive, leading to the deactivation of entire subtrees. This process converts dynamic, input-dependent routing into stable, data-dependent structural sparsity.

Figure~\ref{fig:auto_acc} illustrates this effect by evaluating validation accuracy under fixed statistical pruning ratios. As training proceeds, models with increasing levels of permanent pruning closely match the performance of their unpruned counterparts, indicating that auto-pruning preserves the most informative computation paths.

\subsection{Effect of Nonlinearity Placement}

For the alternative $\mathrm{FFF}^{\mathrm{post}}$ formulation,
\begin{equation}
Y = \mathrm{GELU}(\mathrm{linear\_out}(M \odot Z)),
\end{equation}
the gradient with respect to $z$ no longer depends directly on $\mathrm{GELU}'(z)$, but on a mixed representation after aggregation. As a result, the sign-dependent suppression mechanism described above is absent, and we do not observe stable auto-pruning behavior empirically. This effect instead leads to a more balanced tree utilization, as discussed and tested in Section~3.

\subsection{Pruning results for LLMs}

\begin{figure}
    \centering
    \includegraphics[width=0.5\linewidth]{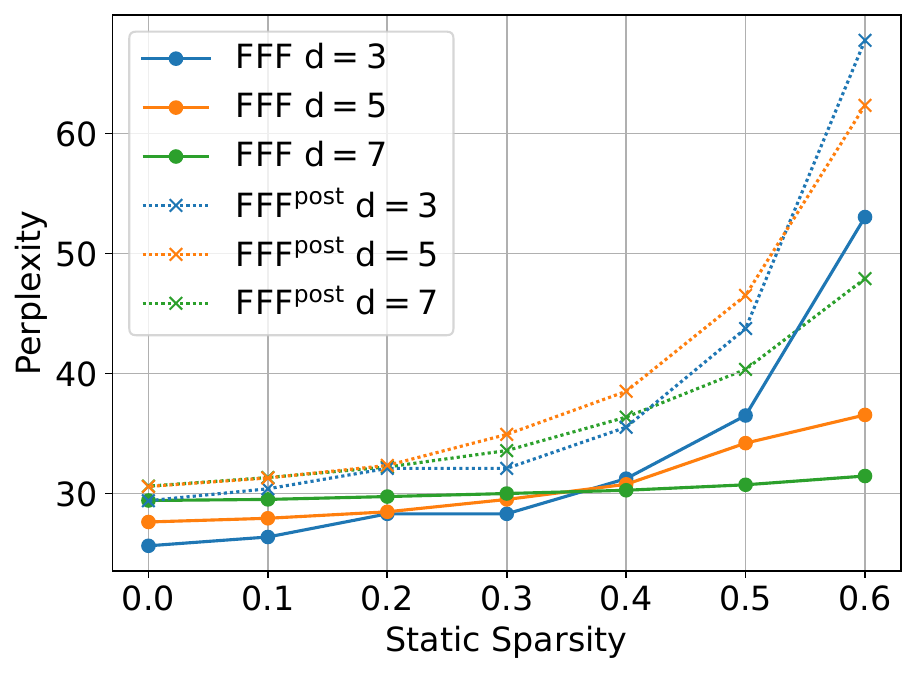}
    \caption{Exploiting static sparsity in OPT-125M model.}
    \label{fig:static-sparsity}
\end{figure}

In Fig.~\ref{fig:static-sparsity} we evaluate the impact of static sparsity on the OPT 125M model. We removed the fractions of least used tree nodes effectively, pruning the model after training. As expected, pruning had higher impact on FFF$^\text{post}$ variant than on FFF. Moreover, pruning was more effective on deeper trees, FFF layers with $d=7$ could be pruned by up to 40\% without significant drop in task performance, but perplexity was overall higher for sparser models.

\subsection{Summary}

Together, these results show that auto-pruning arises from a specific interaction between hard binary routing and the asymmetric GELU nonlinearity in the pre-activation design. This mechanism naturally transforms dynamic routing into stable structural sparsity during training, without explicit pruning rules or auxiliary losses.

\section{Implementation and Training Details}
\label{app:impl}

All experiments were conducted on a multi-node GPU cluster equipped with NVIDIA A100 GPUs. 
For the OPT-125M experiments, we used multi compute nodes, each containing 4 GPUs. 
Training was implemented using the Hugging Face Transformers framework with DeepSpeed integration and the Accelerate library for distributed execution and mixed-precision training.

We followed a Chinchilla-style training paradigm, where training proceeds for a single epoch while controlling the total number of tokens processed rather than training to convergence over multiple epochs. 
The total number of tokens per optimization step was fixed to approximately $5\times10^{5}$ tokens for 125 million paramter model , $1\times10^{6}$ tokens for 1 billion  parameter model, and $2\times10^{6}$ tokens for 3 and 7 billion paramter model. 
Per-device batch size and gradient accumulation steps were chosen accordingly to reach this target token budget across all GPUs.

Optimization was performed using AdamW with fused CUDA kernels, cosine learning rate scheduling with a linear warmup phase, and gradient clipping with a maximum norm of 1.0. 
Training was conducted in FP16 precision. 
No maximum number of steps was imposed, and training was terminated after completing one full pass over the dataset.

For simplicity, the analysis is presented using the masked formulation of the FFF layer, which is functionally equivalent to the sequential tree traversal version described in the main text.

\paragraph{Dataset} The SlimPajama \cite{cerebras2023slimpajama} dataset was used for training. 
The dataset was tokenized offline prior to training using a GPT-style tokenizer. 
Tokenized shards were randomly sampled across runs using a fixed random seed to ensure reproducibility while maintaining stochasticity in shard selection. 
All tokenization and sharding were performed before training to eliminate runtime preprocessing overhead.

\subsection{Hyperparameters}

Table~\ref{tab:hparams-opt125m} summarizes the hyperparameters and training configuration used for the OPT-125M experiments.
Unless stated otherwise, all values were kept fixed across runs for this model size.
Method-specific architectural details are described in the main paper, while this appendix focuses on reproducibility-relevant training and optimization settings.

\begin{table}[H]
\centering
\caption{Hyperparameters and training configuration for OPT-125M.}
\label{tab:hparams-opt125m}
\small
\begin{tabular}{lp{0.6\linewidth}}
\toprule
\textbf{Hyperparameter} & \textbf{Value} \\
\midrule

\multicolumn{2}{l}{\textbf{Model}} \\
\midrule
Base model & \texttt{facebook/opt-125m} \\
Initialization & From scratch \\
Sequence length & 1024 \\
Tokenizer & GPT-style tokenizer (offline tokenization) \\

\midrule
\multicolumn{2}{l}{\textbf{Dataset}} \\
\midrule
Dataset & SlimPajama \\
Tokenization & Pre-tokenized before training \\
Sampling & Random shard sampling with fixed seed \\
Target training tokens & $2.5\times10^{9}$ \\

\midrule
\multicolumn{2}{l}{\textbf{Optimization}} \\
\midrule
Optimizer & AdamW (fused) \\
Learning rate & $1\times10^{-4}$ \\
Scheduler & Cosine \\
Warmup ratio & 0.03 \\
Adam $\beta_1$ / $\beta_2$ & 0.9 / 0.95 \\
Adam $\epsilon$ & $1\times10^{-8}$ \\
Weight decay & 0.01 \\
Gradient clipping & 1.0 \\

\midrule
\multicolumn{2}{l}{\textbf{Training}} \\
\midrule
Epochs & 1 \\
Per-device batch size & 16 \\
Gradient accumulation steps & 4 \\
Precision & FP16 \\
Effective tokens per step & $\approx5\times10^{5}$ \\
Random seed & 42 \\

\midrule
\multicolumn{2}{l}{\textbf{Distributed setup}} \\
\midrule
GPUs & 32 NVIDIA A100 \\
Nodes $\times$ GPUs per node & 8 $\times$ 4 \\
Parallelism & Data parallel (DeepSpeed) \\
Framework & Hugging Face Transformers + Accelerate \\

\bottomrule
\end{tabular}
\end{table}

\section{Inductive Biases for Heavy-Tailed Distributions in Binary Trees}
\label{app:tree-prior-analysis}

\begin{figure}
    \centering
    \includegraphics[width=0.5\linewidth]{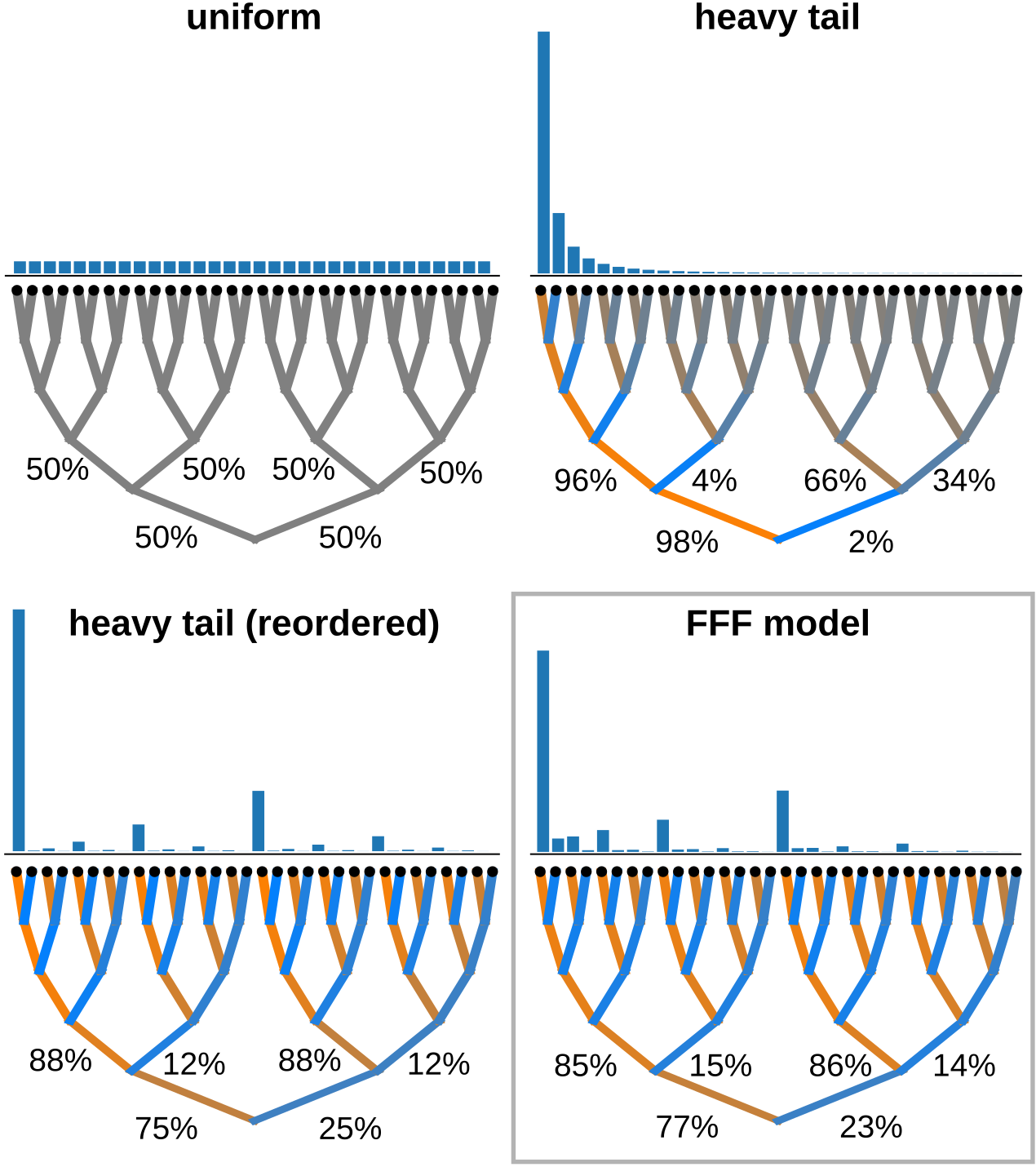}
    \caption{Binary trees with different branch weights and corresponding induced path distributions.}
    \label{fig:tree-prior-analysis}
\end{figure}

It can be easily seen that a binary tree can be constructed to realize any distribution over paths. This can be done by starting with any target distribution, assigning this to leaves, re-normalizing, and iteratively summing probabilities of child nodes until the root is reached. Figure ~\ref{fig:tree-prior-analysis} illustrates this. 4 different binary trees with $d=5$ are shown, a uniform distribution, an approximate heavy tailed distribution with finite support and the same heavy tailed distribution with re-ordered leaves (indices ordered breadth first instead of depth first). The heavy tailed distributions are given here by Pareto distributions with $\alpha=2.0$. Branch weights are indicated by color and labels for first 2 levels. The bottom right tree is generated based on the mean path utilization from our imbalanced model as in Fig.~\ref{fig:tree_util_2}.
It can be clearly seen that the branch weights are more similar to the heavy tailed model (bottom left), and that this model is best compatible with the imbalanced tree structure (left child preferred over right). Heavy-tailed distributions have been identified as a powerful inductive bias that could be responsible for some of the emergent properties of LLMs \cite{chan2022data}. Our results suggest that FFFs have an inductive bias built in, that skews path utilization to mimic such heavy tailed distributions.

\section{Visualization of Space Partitioning}
\label{app:space-partition}

To illustrate the structured behavior of tree-based routing, we analyze an FFF layer on a simple 2D toy task. The input is a point $(x_1, x_2)\in\mathbb{R}^2$ and the target is a binary label defined by a checkered-field (checkerboard) pattern over the input space. This setting allows us to directly visualize how hierarchical routing partitions the input domain.

\begin{figure}[h]
    \centering
    \includegraphics[width=\linewidth]{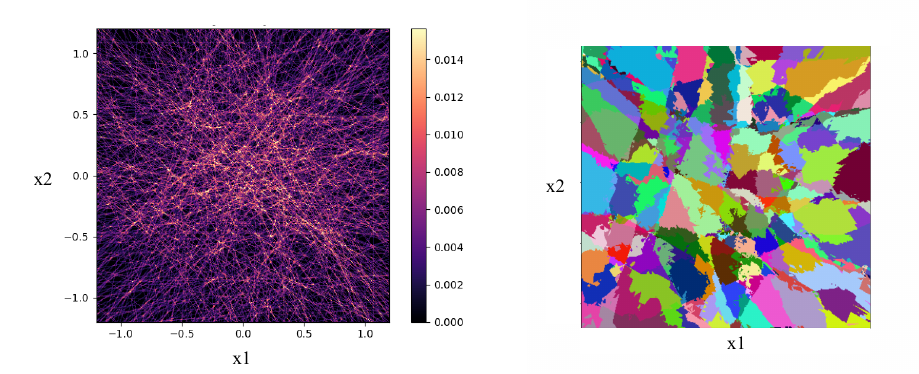}
    \caption{Visualization of FFF-induced space partitioning on a 2D checkered-field classification task. 
    Left: routing-induced decision boundaries mapped back to the input space; each split corresponds to a node-level routing decision, and the combined boundaries show how the forest partitions the plane into routed regions. 
    Right: PCA projection of representations grouped by routing outcomes; colors indicate distinct routed regions (sets of active paths) that contribute differently to the output.}
    \label{fig:toy_partition}
\end{figure}

Figure~\ref{fig:toy_partition} provides two complementary views of the learned partitioning. The left panel visualizes the decision boundaries induced by tree traversal, mapped back into the original input space. Each line corresponds to a routing split at some node, and together the boundaries show how the forest decomposes the 2D plane into a collection of routed regions. The right panel visualizes the resulting routed regions in representation space: we group inputs by their routing outcome (i.e., the set of active paths) and project the corresponding representations using PCA. Regions are color-coded, highlighting that different parts of the input space are consistently assigned to different routing patterns that contribute differently to the final output.

\section{Qualitative Inference Examples}
\label{app:qualitative_examples}

This appendix presents representative prompt--output examples generated by the OPT-1.3B using FFF layers of depth of 4.
The examples illustrate typical behaviors observed during inference, including partial instruction
following, verbosity issues, and occasional semantic errors.
All outputs are shown verbatim and have not been edited for correctness or style.

\textbf{Prompt}
\begin{verbatim}
Complete the following paragraph in a neutral academic tone:

Large language models can be evaluated using both quantitative metrics
and qualitative analysis. In this study, we focus on
\end{verbatim}

\textbf{Model Output 1.3b d=4}
\begin{verbatim}
The quality of the model training data used to train it;
How well the model performs in real-world scenarios;
How well the model adapts to changing conditions; and
Its scalability.

The metric used in this paper is called Quality Assurance Score (QAS).
This metric describes how good an approach, model or strategy really is,
based on both the data itself as well as on a large body of experience
with it.

For example, if there are many available resources such as text
collections that contain a lot of content, then the QAS score will be
\end{verbatim}

\textbf{Model Output 1.3 d=6}
\begin{verbatim}
qualitative analysis of language models. We employ the methods of
meta-analysis to identify models that have the highest quality and
which best represent our domain and dataset.

We first introduce the concept of a language model using the example
of an agent learning the English language through a text-to-speech
model. A number of factors are considered when deciding what to
compare the model with. These include the level of expertise in the
user community, the complexity of the task, the availability of
source code, and other characteristics. For the given set of factors,
we rank each of the models against all available others. Then, we
evaluate the differences between them using the metric “quality”,
which is a measure of accuracy, recall, and performance relative to its
predecessors. Finally, we compare the results from these methods
for the same classifier (or for multiple classes) to see how they
fit.

Our approach has the potential to be a powerful approach to
benchmarking language models at scale. It allows us to identify the
best models as well as their weaknesses. This would allow us to improve
the existing models by incorporating new features or incorporating
\end{verbatim}

---

\subsection{Instruction Following with Constraints}

\textbf{Prompt}
\begin{verbatim}
Write a single paragraph (120–160 words) explaining what "temperature"
does in text generation. Do not use bullet points. Include exactly one
example sentence that shows a difference between low and high temperature.
\end{verbatim}

\textbf{Model Output 1.3 d=4}
\begin{verbatim}
The examples should be from the same paper, or from a previous course
you have taken. You must include an image to accompany your paragraph.

To see the attached sample of a paper written in response to this prompt,
click on the link below:

The main function of the body paragraphs is to support the narrative flow
of your paper. The body paragraphs are also intended as a way for readers
to see your ideas and learn more about your topic. This will help them
understand how you think and why you believe the things you do. Do not
over-write your body paragraphs. They should be short and concise.
Include two photographs in your body paragraphs. For the first
photograph, choose a photo that is representative of the topic. For the
second photograph, select a photo that has something interesting about
it.

2. Explain one thing that can be done with text generated by the
software. Show how you would implement this feature. Discuss the pros
and cons of this feature. How long would it take to implement this
feature?

3. Write a report/paper using text
\end{verbatim}

\textbf{Model Output 1.3 d=6}
\begin{verbatim}
Include the URL of the page where you found this information.

This assignment may be used in any course in which students are required to write
a report or some type of written expression. See the list of course assignments for
more details on course requirements. You must obtain permission from your professor
before using this assignment.

If you have trouble writing this assignment, consider looking at examples before writing
your own. This assignment is designed to help you learn how to write well, but it does
not replace the time and effort involved in mastering skills such as grammar, spelling,
and punctuation.
\end{verbatim}

---

\subsection{Lexical Ambiguity Resolution}

\textbf{Prompt}
\begin{verbatim}
The term "bank" can have multiple meanings. Provide two interpretations
of the sentence below and explain which is more likely given the context.

Sentence: "I went to the bank after work."
Context: "I needed to deposit my paycheck."
\end{verbatim}

\textbf{Model Output 1.3b d=4}
\begin{verbatim}
I went to the bank after work.
I did not need to deposit my paycheck.

The first meaning is a negative one, because we know that after you
deposit your paycheck at the bank (or any other financial institution
for that matter), you will leave the money there for the week while you
are working. Therefore, you would never go to the bank after you had
deposited it. The second meaning of "went to the bank" is "went to a
place where someone else had deposited a check and he or she wanted to
put their own money in." In this situation, the person was depositing
his/her own money, but the bank was already holding and using the money.
Thus, we could say that the person went to the bank AFTER he/she had
deposited the money (after work). So the second definition is incorrect.

Solution:

This question can be answered with the following example.

Sentence: "I walked into the library this afternoon looking for a book
on the psychology of addiction."
Context: "I needed a book about addiction."

The person walking into the library is asking whether or not they have a
chance of finding something useful that they do not know about. There
\end{verbatim}

\textbf{Model Output 1.3 d=6}
\begin{verbatim}
Bank: "You deposited your paycheck?" [Note that this could have been a cashier's check,
but
since we are not dealing with cash, we will assume it was a paycheque.]
[Note that "the bank" could also mean a financial institution such as a brokerage firm
or credit union.]

Sentence: "I told him I would be there in fifteen minutes."
Context: "I had just left the store where I bought my groceries and was in the car
parking lot waiting for the next bus."
Bank: "That sounds like a long time!"
[This could be a cashier's check, but since we are not dealing with cash, we will assume 
it was a
paycheque.]

Sentence: "When I got home, I discovered that my husband had cooked dinner for me."
Context: "He must have decided that it was time for us to eat dinner together."
Bank: "It is possible that you forgot to tell him that you were going to the store
after work."
[This could be a cashier's check, but since we are not dealing with cash, we will assume 
it was a
paycheque.]
\end{verbatim}
---

\subsection{Numerical Reasoning}

\textbf{Prompt}
\begin{verbatim}
A model produces 180 tokens per second on GPU. How long does it take to
generate 900 tokens? Show the calculation and give the final answer in
seconds.
\end{verbatim}

\textbf{Model Output 1.3 d=4}
\begin{verbatim}
If this question can be reworded to fit the rules in the help center,
please edit the question.

3 Answers
3

You're talking about a GPU that is not really suited for such kind of
calculations. So, assuming we have a GPU that can do 10^9 operations per
second, our best guess would be to add up all the number of operations
(the numbers you saw above) with the amount of time they took (the time
you were asking). I'm afraid I have no idea how the output value of the
GPU will be used. So, you might want to specify what purpose you're
planning to use its outputs. Maybe you could share your project code
here: 
\end{verbatim}

\textbf{Model Output 1.3 d=6}
\begin{verbatim}
There are 4 types of GPUs used for crypto mining: the graphics card, the ASIC 
(application-specific integrated circuit), the TPU (thin-client processor) and the MCU 
(microcontroller unit). Show how they work.

In this lesson you’ll learn how to generate bitcoin on a dedicated PC running Linux. 
In addition
to showing you how to mine bitcoin with your own hardware, you'll also see some easy 
methods to get started with cloud mining for bitcoin and other cryptocurrencies, such as 
Ethereum, Litecoin and Dash.

You’ll start by downloading and installing Bitcoin Core on your computer, then you’ll 
configure the wallet and set up the mining program with it. You can even choose to either
run the client directly or use the “lite” client, which is just a faster version of the full 
client that comes with the blockchain. The miner will then start to download and verify
transactions from the network. At each block you’ll be given an incentive 
(either a small amount of bitcoins or a percentage of the fee paid out) to keep mining on. 
If you receive no payment at all, you will continue mining until the reward for finding 
the next block is bigger
\end{verbatim}


\end{document}